# Steam Turbine Anomaly Detection: An Unsupervised Learning Approach Using Enhanced Long Short-Term Memory Variational Autoencoder

Weiming Xu and Peng Zhang[*]

*Department of Mechanical Engineering, City University of Hong Kong, Kowloon Tong, Kowloon, 999077, Hong Kong*

## Abstract

As core thermal power generation equipment, steam turbines incur significant expenses and adverse effects on operation when facing interruptions like downtime, maintenance, and damage. Accurate anomaly detection is the prerequisite for ensuring the safe and stable operation of steam turbines. However, challenges in steam turbine anomaly detection, including inherent anomalies, lack of temporal information analysis, and high-dimensional data complexity, limit the effectiveness of existing methods. To address these challenges, we proposed an Enhanced Long Short-Term Memory Variational Autoencoder using Deep Advanced Features and Gaussian Mixture Model (ELSTMVAE-DAF-GMM) for precise unsupervised anomaly detection in unlabeled datasets. Specifically, LSTMVAE, integrating LSTM with VAE, was used to project high-dimensional time-series data to a low-dimensional phase space. The Deep Autoencoder-Local Outlier Factor (DAE-LOF) sample selection mechanism was used to eliminate inherent anomalies during training, further improving the model's precision and reliability. The novel deep advanced features (DAF) hybridize latent embeddings and reconstruction discrepancies from the LSTMVAE model and provide a more comprehensive data representation within a continuous and structured phase space, significantly enhancing anomaly detection by synergizing temporal dynamics with data pattern variations. These DAF were incorporated into GMM to ensure robust and effective unsupervised anomaly detection. We utilized real operating data from industry steam turbines and conducted both comparison and ablation experiments, demonstrating superior anomaly detection outcomes characterized by high accuracy and minimal false alarm rates compared with existing methods.

---

[*] Corresponding author
 E-mail address: penzhang@cityu.edu.hk, Tel: (852)34429561.

# Introduction

Thermal power generation is a cornerstone of global electricity production, playing a pivotal role in driving industrial productivity, fostering socio-economic growth, and profoundly impacting daily livelihoods [1]. As the core equipment of the energy ecosystem, steam turbines are an indispensable and intricate instrument, incurring substantial expenses when facing downtime, maintenance, and damage. The intricate operational demands and frequent high-load exposures in thermal power generation systems lead to random and diverse turbine failures, including thermal stress, material degradation, lubrication issues, and manufacturing defects. This poses significant challenges for fault diagnosis research, as accumulating adequate historical fault data with precise labeling of all potential fault types is both costly and complex [2].

Unsupervised anomaly detection technology offers a promising solution. It circumvents the need for exhaustive historical labeled datasets across all fault categories, yet it holds profound implications for enhancing the reliability and stability of thermal power generation systems [3]. Anomalies in steam turbines manifest as either abrupt changes [4], often caused by sudden operational shifts, or gradual changes [5], resulting from factors such as fouling, cylinder wall wear, and blade deformation [6-8]. Typically, these abnormalities are detected by assessing whether the patterns of relevant sensor data (such as pressure, temperature, vibration, etc.) conform to expected behavior [9].

Anomaly detection methods can be broadly categorized into analytical model-based and data-driven approaches. While analytical methods rely on constructing precise mathematical-physics models, this becomes increasingly challenging as modern machines grow in complexity. Data-driven methods leverage advancements in sensor technology and computing power to detect anomalies [10, 11]. Unsupervised steam turbine anomaly detection is critical for promptly identifying deviations from normal operation, particularly in the early stages of malfunction [12]. A notable advantage is their reduced dependence on labeled data and specialized expertise, making them particularly valuable for early-stage malfunction detection in steam turbines.

Among data-driven unsupervised approaches, autoencoder-based methods have gained prominence for their robustness across various anomaly detection tasks. These methods typically follow two main strategies. The first involves leveraging reconstruction errors between original and reconstructed data to identify anomalies. Notably, Xia et al. [13] demonstrated the efficacy of this approach in distinguishing inliers from outliers across image datasets. The second strategy employs autoencoders for nonlinear dimensionality reduction, enabling the extraction of meaningful low-dimensional representations from high-dimensional data. For instance, Dairi et al. [14] enhanced anomaly detection in urban environments by integrating latent features from Deep Stacked Autoencoders with k-nearest neighbor techniques. Although reconstruction-based methods are effective, they often rely on empirically set thresholds, which can be challenging for rare fault detection. In contrast, Variational Autoencoders (VAE) [15], offer enhanced generalization and interpretability through structured regularization of the latent space [16]. This study leverages the VAE's strengths as a core component in developing a novel reduced-order model.

Given the inherent temporal dynamics in steam turbine operation datasets, especially those involving gradual degradation faults, incorporating temporal information into anomaly detection models is both natural and necessary. Recent studies have explored various approaches to integrating temporal dynamics. Lei et al. [17] used a dual-attention autoencoder with GRU cells for improved alarm timing and stability. Hu et al. [18] used a Long Short-Term Memory-based (LSTM-based) autoencoder to achieve real-time detection of abnormalities. Liu et al. [19] combined stacked denoising autoencoders with LSTM networks to enhance forecasting for a 1000 MW USC unit. To address these challenges, we introduce the chrono-attention reduced-order model, Long Short-Term Memory Variational Autoencoder (LSTMVE), a novel framework integrating LSTM networks with VAE. This approach captures essential spatial-temporal characteristics within a low-dimensional latent space, enhancing continuity and probabilistic interpretation. By preserving crucial temporal dynamics, the LSTMVAE model enables more nuanced detection of anomalies, particularly those arising from gradual degradation. Unlike traditional methods that rely solely on reconstruction errors, this framework offers a more sophisticated solution for anomaly detection in complex steam turbine systems.

In the complex operational environment of steam turbines, data credibility is often compromised by sensor inaccuracies, environmental interferences, and mechanical degradation. Consequently, the steam turbine operation data frequently exhibit intrinsic noise and anomalies, which can significantly challenge the training and efficacy of anomaly detection algorithms. To address this, we proposed a Deep Autoencoder-Local Outlier Factor (DAE-LOF) sample selection mechanism data screening mechanism that integrates the high-dimensional analytical power of deep autoencoders with the density-based Local Outlier Factor (LOF) [20] algorithm. This multi-sensor fusion approach aims to enhance LSTMVAE's performance by effectively mitigating the impact of intrinsic noise and anomalies during training, thereby enhancing the model's discriminative ability and focusing on core healthy data [21, 22].

Building upon prior research, the current research proposes an Enhanced Long Short-Term Memory Variational Autoencoder using Deep Advanced Features and Gaussian Mixture Model (ELSTMVAE-DAF-GMM), an innovative architectural framework built for improving unsupervised anomaly detection accuracy in steam turbines. The dataset comprises time series signals from industrial steam turbine systems, recorded under varied operational conditions, including both normal and abnormal states. Each mode has been rigorously validated by industry experts, ensuring that the data accurately reflects the full range of operational phenomena in steam turbines. The ELSTMVAE model combines a DAE-LOF sample selection mechanism with the LSTMVAE model, creating a robust framework that fuses temporal dynamics with reconstruction discrepancies. Furthermore, by integrating latent embeddings that capture temporal dynamics with reconstruction discrepancies, the new deep advanced features (DAF) from the LSTMVAE model deliver a more nuanced data representation. Finally, incorporating these deep advanced features into a Gaussian Mixture Model ensures robust and effective unsupervised anomaly detection, significantly enhancing the model's capability to identify anomalies in complex steam turbine operations.

# 1. Deep Learning Model Architecture and Dataset Characterization

As discussed in the Introduction, detecting anomalies in high-dimensional and complex steam turbine systems presents significant challenges to traditional methods. The primary difficulty lies in extracting meaningful features from extensive sensor networks, which involves navigating the intricate spatial coupling of physical interactions and dynamic temporal changes. Additionally, it is crucial to mitigate the impact of inherent noise and anomalies arising from the complex environment and system variations. These challenges are paramount in accurately identifying the operational state patterns of steam turbines and are overcome in the present study by adopting the following methodology.

*1.1 Enhanced Long short-term memory variational autoencoder design*

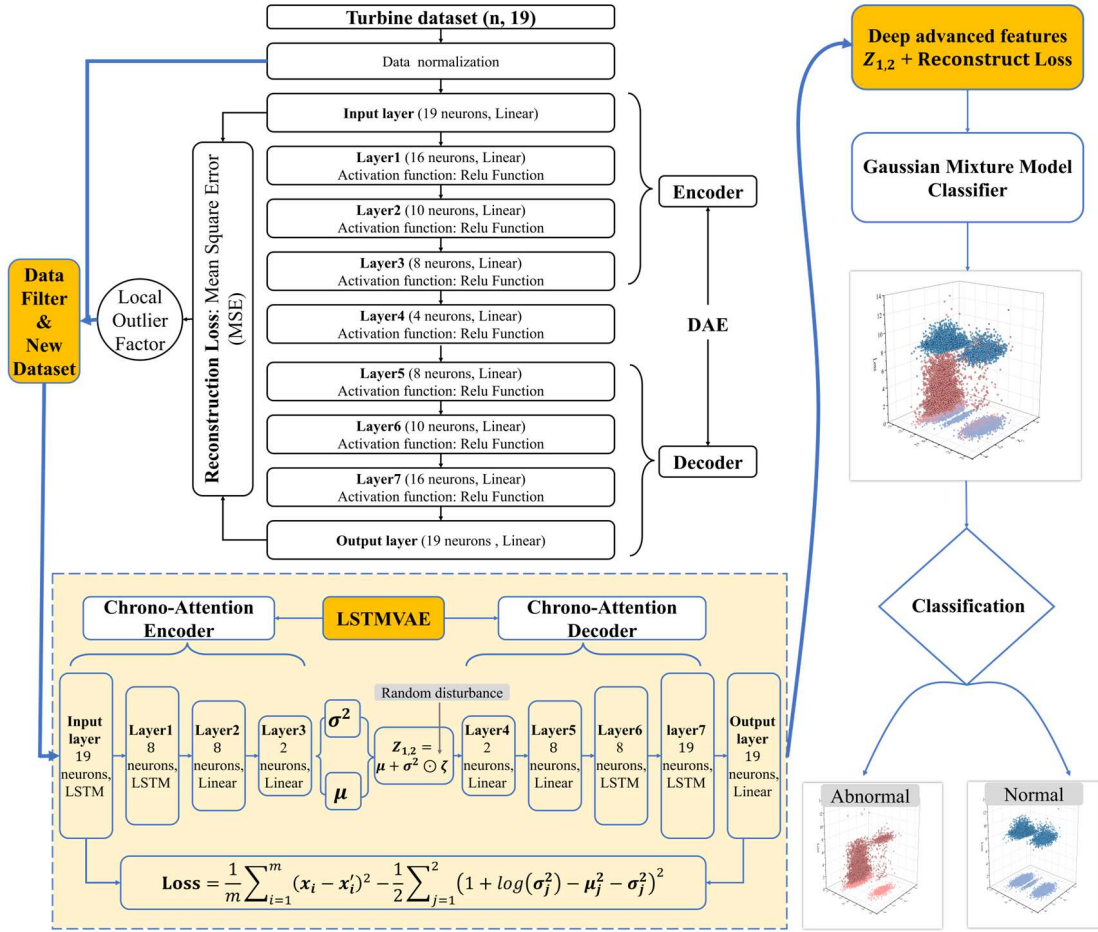

Fig. 1. The structure of the designed ELSTMVAE-DAF-GMM method consists of three modules: a DAE-LOF sample selection mechanism, a neural network model of LSTMVAE, and a flowchart of the GMM unsupervised classifier.

An Enhanced Long Short-Term Memory Variational Autoencoder using Deep Advanced Features and Gaussian Mixture Model (ELSTMVAE-DAF-GMM) novel deep learning approach is for the first time proposed to improve unsupervised anomaly detection accuracy in the steam turbine. Fig. 1 depicts a detailed design of the framework, which is composed of two components: an ELSTMVAE model integrating the LSTMVAE with a Deep Autoencoder-Local Outlier Factor (DAE-LOF) sample selection mechanism, and a GMM classifier.

The ELSTMVAE is an innovative methodology introduced in this study to enhance the dimensionality reduction capability of the VAE model by combining LSTM and the DAE-LOF sample selection mechanism. To address the challenge of disruption of intrinsic noise and anomalies during training, a multi-sensor fusion sample selection mechanism DAE-LOF is proposed by combining DAE with LOF algorithms, focusing on identifying and mitigating core healthy data from the training set correlated features within steam turbine sensor data and enhance the robustness of the reduced-order model. To train the deep autoencoder within the sample selection algorithm, we adopted a four-layer configuration for both the encoder and decoder. The encoder comprises two initial linear layers with 19 and 16 neurons, followed by three subsequent layers with 10-8-4 neurons, respectively. Similarly, the decoder consists of five linear layers with 4-8-10-16-19 neurons. For the steam turbine dataset, we utilized the LOF algorithm to identify samples with significant deviations in their reconstruction errors compared to the local error clusters. These outliers, denoted by sample h, are then removed from the dataset. Meanwhile, sample $q$, which exhibits the closest reconstruction error to that of normal samples, is selected as a reference for normal behavior. This process enables the construction of a refined training dataset tailored for subsequent analysis using the LSTMVAE algorithm. The contamination calculation formula

$$C = \frac{h}{h+q} \times 100\% = 20\% \tag{1}$$

serves as a controllable parameter in this methodology and will be discussed shortly in the following section. Further elaboration on the mathematical theorem of the LOF algorithm can be found in the first section of the Supporting Material. This meticulous data refinement process ensures the removal of disruptive outliers during the training process, thereby bolstering the overall resilience and performance of the proposed models. The results in Section 3.2 of the industrial steam turbine contrast experiment highlight the significance of employing advanced sample selection techniques to enhance the reliability of predictive models within critical industrial environments.

The LSTMVAE model mainly consists of two integral components: VAE and two-layer LSTM. VAE, a powerful tool for anomaly detection based on deep learning, has found widespread applications in steam turbine monitoring. Comprising encoder and decoder modules constructed with multiple neural network layers, VAE can efficiently map information from high-dimensional monitoring sensor data to a compact low-dimensional phase space with continuity and probabilistic interpretation, as shown in Fig. 2.

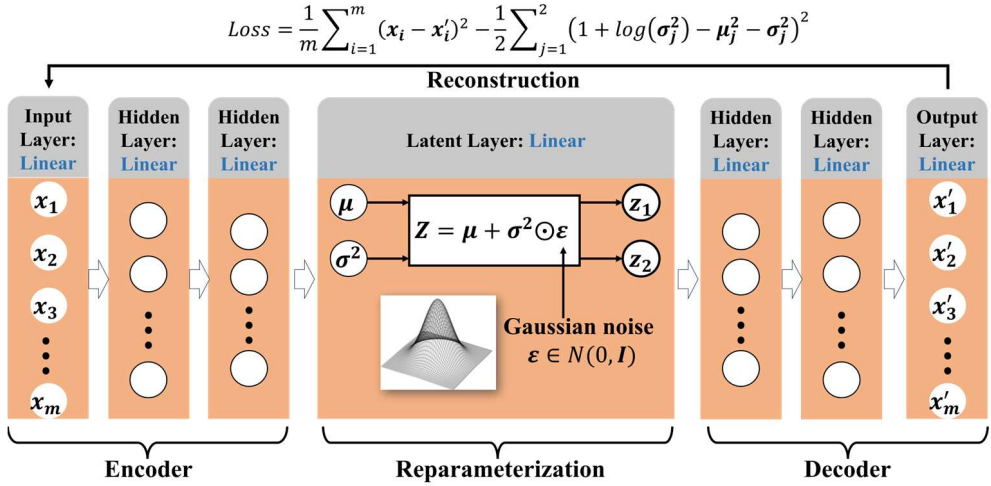

Fig. 2. The structure of a conventional variational autoencoder consists of three modules: the encoder, reparameterization, and decoder.

In the encoder, the steam turbine data are nonlinearly mapped to the latent variable through the multiple linear neural network layers with decreasing numbers of neurons, ultimately producing a lower-dimensional latent space. The relationship between the input data $X = \{x_1, x_2, \cdots, x_n\}$ and the output of the encoder $Y = \{\mu, \sigma^2\}$ can be expressed as

$$\{\boldsymbol{\mu}, log\boldsymbol{\sigma}^2\} = f_\Theta(\boldsymbol{X}; \boldsymbol{\theta_\mu}, \boldsymbol{\theta_\sigma}) \qquad (2)$$

where $\boldsymbol{\mu}$ and $\boldsymbol{\sigma}^2$ denote the parameters outputted by the encoder, using the log variance $log\boldsymbol{\sigma}^2$ instead of the variance $\boldsymbol{\sigma}^2$ provides numerical stability and facilitates optimization. The encoder, represented by the neural network $f_\Theta$, is defined by the parameter set $\Theta$, encompassing all network weights and biases. Specifically, $\boldsymbol{\theta_\mu}$ and $\boldsymbol{\theta_\sigma}$ are the weight and bias of the encoder that are used to compute $\boldsymbol{\mu}$ and $log\boldsymbol{\sigma}^2$, respectively.

In the latent layer, to handle the challenge of sampling from a Gaussian distribution during gradient-based optimization, VAE incorporates a reparameterization trick to enhance gradient optimization. This technique is depicted by

$$\boldsymbol{Z} = \boldsymbol{\mu} + \boldsymbol{\sigma}^2 \odot \boldsymbol{\varepsilon} \qquad (3)$$

which introduces stochasticity into the latent variable $\boldsymbol{Z} = (z_1, z_2, \cdots, z_m)$ as a sample drawn from a distribution $N(\boldsymbol{\mu}, \boldsymbol{\sigma}^2)$, where $\boldsymbol{\mu}$ and $\boldsymbol{\sigma}^2$ are learned parameters. Direct sampling of the stochastic variable from this Gaussian distribution can complicate gradient computation during backpropagation, thereby hindering effective parameter optimization [23]. To address this, random sampling from a Gaussian noise distribution $\boldsymbol{\varepsilon} \sim N(\boldsymbol{0}, \boldsymbol{I})$ is employed, where $\odot$ denotes the Hadamard product.

In the decoder, the latent variable $\boldsymbol{Z}$ is reconstructed to decoder output $\boldsymbol{X'} = (x_1', x_2, \cdots x_n')$ through the multiple linear neural network layers with increasing numbers of neurons to match the dimension of the input data, as given by

$$\boldsymbol{X'} = f_\Phi(\boldsymbol{Z}; \boldsymbol{\phi_Z}) \qquad (4)$$

where $\boldsymbol{X'}$ is the output vector of the decoder, $f_\Phi$ is a neural network of the encoder. The parameter set $\Phi$ includes all the weights and biases that define the decoder's structure and functionality. $\boldsymbol{\phi_Z}$ are the weight and bias of the decoder that are used to compute $\boldsymbol{X'}$.

To learn the temporal dynamic behaviors inherent in complex steam turbine operation datasets, particularly within the datasets characterized by gradual degradation fault occurrences, we have implemented a two-layer LSTM network, as depicted in Fig.

3(d). This approach replaces the fully connected layers commonly used in conventional variational autoencoders, as illustrated in the encoder in Fig. 3(b). This modification facilitates the development of both the chrono-attention encoder and the chrono-attention decoder by integrating the two-layer LSTM network with the VAE, enhancing the model's capability to capture and analyze temporal dynamics.

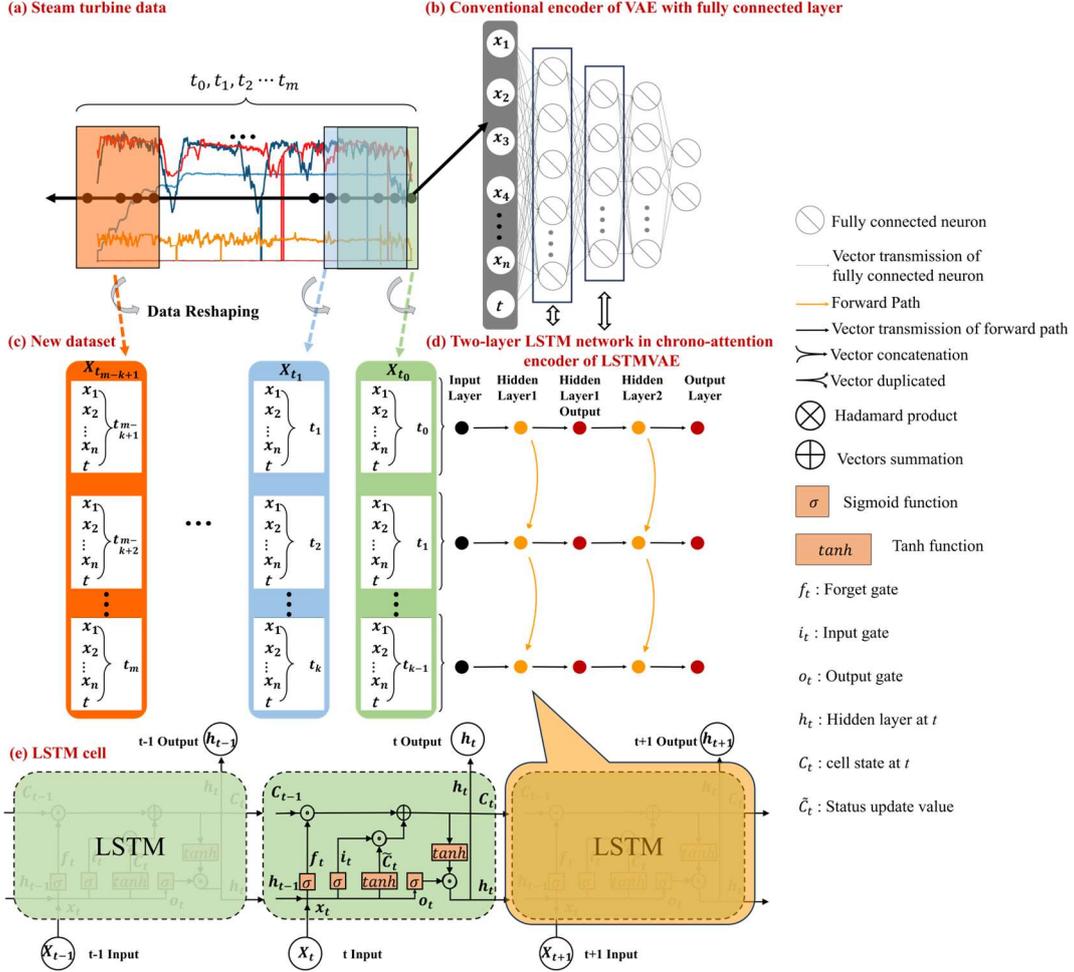

Fig. 3. Comparison of the chrono-attention encoder of LSTMVAE with the conventional encoder of VAE. (a) Time series data from the steam turbine. (b) Conventional VAE encoder with fully connected layers. (c) New dataset after data reshaping. (d) Two-layer LSTM network in the chrono-attention encoder of LSTMVAE. (e) Long Short-Term Memory (LSTM) cell structure.

LSTM, a specialized form of Recurrent Neural Networks (RNNs), was initially proposed by Hochreiter et al. [24] as a pivotal solution to the challenges faced by conventional neural networks, which are limited in their capacity to effectively capture temporal dependencies within time series data. As depicted in Fig. 3(e), the core of the LSTM neural cell lies in its three gating mechanisms, each playing a crucial role in regulating information flow within the network. Among these gating mechanisms, the forget gate mechanism stands out as a fundamental component, allowing the network to selectively retain or discard information based on its relevance to the current context. Mathematically, the forget gate's operation is expressed by

$$F_t = \sigma(W_f \cdot [h_{t-1}, x_t] + b_f) \tag{5}$$

where $F_t$ represents the output of the forget gate, indicating the proportion of information slated for removal; $W_f$ and $b_f$ correspond to the weight and bias matrix associated with the forget gate, respectively; $[h_{t-1}, x_t]$ represents the concatenation of the previous time step's hidden state $h_{t-1}$ and the current time step's input $x_t$, forming a single vector.; $\sigma$ is typically an activation function.

The input gate and output gate mechanisms are integral to LSTM's functioning. The input gate mechanism, denoted by $I_t$, governs the inflow of new information into the cell state and is calculated by

$$I_t = \sigma(W_i \cdot [h_{t-1}, x_t] + b_i) \tag{6}$$

determines the relevance of new information for inclusion in the cell state, with $W_i$ and $b_i$ representing the associated weight matrix and bias term, respectively. The output gate mechanism, represented by $O_t$, controls the flow of information from the cell state to the hidden state and is calculated by

$$O_t = \sigma(W_o \cdot [h_{t-1}, x_t] + b_o) \tag{7}$$

where $O_t$ indicates the proportion of information to be propagated to the hidden state, with $W_o$ and $b_o$ as the corresponding weight matrix and bias term, respectively.

Additionally, $C_t$ in LSTM networks stands for the cell state, which serves to store and transmit memory information. The relationship between the $C_t$ and $C_{t-1}$ of LSTM neurons is given by

$$C_t = C_{t-1} \odot F_t + I_t \odot [tanh(W_C X_t + W_{hC} h_{t-1} + b_C)] \tag{8}$$

where $W_C$, $W_{hC}$ and $b_C$ represent the weights and bias for $C_t$, respectively. Through gate mechanisms, the network can selectively retain or forget past information to adapt to anomaly detection tasks. The updating and regulation of this memory cell state assist LSTM neural networks in effectively learning and remembering long-term dependencies, playing a crucial role in sequence modeling tasks. These gating mechanisms collectively endow LSTM networks with the capability to effectively capture and utilize long-term dependencies in sequential data, making them a cornerstone in various fields such as anomaly detection in steam turbines [25], time series analysis [26], and mode recognition [27].

The depiction on Fig. 3(d) illustrates the structure of a two-layer LSTM network, comprising two LSTM layers stacked on top of each other. The key characteristic of a two-layer LSTM is its ability to enable hierarchical feature extraction. Each layer processes sequential data and extracts features before passing them to the next layer. Typically, the first LSTM layer captures lower-level features from the input sequence, while the second layer can further extract higher-level abstract features. In summary, a two-layer LSTM offers superior feature extraction capabilities, enhanced representation learning, and greater model capacity compared to its single-layer counterpart, rendering it an invaluable asset for various tasks involving sequential data processing.

The training objective of LSTMVAE is multifaceted. Primarily, it seeks to minimize reconstruction errors between the decoder's output $X'$ and the original input data $X$. Concurrently, it employs a KL divergence term to enforce the alignment between the distribution of latent representations and a predefined prior distribution, typically a Gaussian. Specifically, we aim to maximize the Evidence Lower Bound (ELBO), which serves as a lower bound on the marginal likelihood

$$\log(P(x)) = E_{z \sim q_\theta}[log p(x|z)] - KL(q_\theta(z|x)||p(z)) \qquad (9)$$

where

$$E_{z \sim q_\theta}[log p(x|z)] = \frac{1}{n}\sum_{i=1}^{n}(x_i - x_i^{''})^2 \qquad (10)$$

represents the reconstruction error between the input $X$ and output $X''$, and

$$KL(q_\theta(z|x)||p(z) = \frac{1}{2}\sum_{j=1}^{2}(1+log(\sigma_j^2)-\mu_j^2-\sigma_j^2) \tag{11}$$

represents the regularization term of the posterior distribution $q_\theta(z|X)$ and the prior distribution $p(z)\sim N(0,I)$. Therefore, the training objective can be represented by the loss function

$$Loss = \frac{1}{m}\sum_{i=1}^{m}(x_i-x_i')^2 - \frac{1}{2}\sum_{j=1}^{d}(1+log(\sigma_j^2)-\mu_j^2-\sigma_j^2) \tag{12}$$

where $\sigma$ and $\mu$ are the mean and standard deviation of the latent space representations, respectively. $m$ and $d$ respectively denote the number of samples and the dimension of latent variables. This dual objective ensures that the LSTMVAE learns both faithful reconstructions and meaningful latent representations. Through weight updates during training, the LSTMVAE iteratively adjusts its parameters to optimize this combined objective.

The behavior of steam turbines exhibits complexity, often involving numerous sensors with diverse impacts such as steam turbine blade wear affecting various operational parameters like efficiency, temperature, and pressure. Both the precise selection of sensors and the direct analysis of high-dimensional data for anomaly detection pose considerable challenges. Utilizing the dimensionality reduction capabilities inherent in LSTMVAE architectures and the extensive feature learning capabilities of neural networks, our proposed methodology endeavors to identify the present operational state of steam turbines. This is accomplished through an exploration of the low-dimensional latent features and reconstruction errors within the original dataset, denoted as Deep Advanced Features (DAF), thereby facilitating the construction of a comprehensive phase space representation.

Both the DAE and LSTMVAE neural network in the ELSTMVAE model are using the same training strategy. The optimization employs the Adam optimization algorithm, and the training method is the minibatch training method with a batch size of 64. The model training process uses an early stopping strategy, that concludes either after 100 iterations with no loss reduction or reaching the pre-defined epoch limit of 20000. These neural networks are coded by using the deep learning structure Pytorch as the backend computation library. To train the neural network proposed in this study, we set five layers in the encoder and decoder. In the encoder, the first two are two-layer LSTM layers with 19 and 8 neurons respectively, followed by a fully connected layer with 8 neurons, and the last layer is a fully connected layer with 2 neurons. For the decoder, there are two two-layer LSTM layers with 8-19 neurons, and a fully connected layer with 19 neurons. The entire code execution takes approximately 4405 seconds to process about 11400 training samples on the NVIDIA Tesla V100SXM2 GPUs at the National Supercomputer Center in Guangzhou. The software libraries and configurations used include Python version 3.8.13, CUDA version 11.6, and Pytorch version 1.12.0+cu121. For additional configurations, please consult the default setup provided by the National Supercomputer Center in Guangzhou.

It is worth mentioning that the dimensionality reduction model in this paper is specifically designed for precise steam turbine anomaly detection. Our dataset shows that the three-dimensional DAF space effectively captures data information, evidenced by minimal reconstruction errors. Distinct phase point distributions for each mode within this space highlight its advantage in mode recognition. These findings, to be further discussed in Section 3, underscore the robust representational power of a three-dimensional phase space for steam turbine data. Furthermore, we noted that some complex systems may require a representation beyond a three-dimensional latent space. This can be achieved by designing an encoder that maps input data to a higher-dimensional latent space without adding essential difficulties to the present approach. For applications involving large datasets, the computational cost of our deep learning approaches can be mitigated through several strategies. These include the use of compression techniques, such as quantization, which reduce the precision of weight matrices (e.g., from 32-bit floating point values to 8-bit integers) with minimal loss of quality [28]. Additionally, learning techniques like distillation can be employed to train smaller models that replicate the performance of larger models, thereby improving accuracy while reducing the parameter count and computational footprint [29]. Automation tools such as Hyper-Parameter Optimization (HPO) and architecture search can also be utilized to optimize hyper-parameters and model architectures, enhancing both accuracy and efficiency and ultimately reducing model size and latency [30].

*1.2 Gaussian Mixture Model*

Acquiring a thoroughly labeled dataset in industrial processes, particularly in complex steam turbine systems, is both costly and challenging. The limited availability of data and the complexity of these systems present significant obstacles. Therefore, the development of effective unsupervised methods for steam turbine anomaly detection is a promising direction. These methods provide crucial insights into the behavior of data patterns within steam turbine systems, especially in contexts where data is sparse and not well understood. In this study, the deep advanced features extracted from the steam turbine dataset are inputted into a Gaussian Mixture Model (GMM) for unsupervised anomaly detection. For the final detection task, we strived for the algorithm to adeptly handle time-series data and demonstrate exceptional detection performance, thereby realizing an intelligent unsupervised anomaly detection method for steam turbines.

GMM [31], a type of clustering method based on machine learning, has gained widespread popularity in mode recognition. GMM is a probability model that consists of *K* sub-distributions, each of which is a single Gaussian model. The probability density function (PDF) of a GMM is formed by a linear combination of individual Gaussian probability density functions. If there are K Gaussian distributions, each is defined by its mean, covariance, and weight. The overall PDF is a weighted sum of these individual Gaussian PDFs, with each Gaussian distribution's contribution determined by its corresponding weight. The expression of the PDF for a GMM with *K* Gaussian components is

$$\boldsymbol{P(X)} = \sum_{i=1}^{k} \omega_i N(X|\mu_i, \Sigma_i) \qquad (13)$$

where $X = (x_1, x_2, \cdots, x_n)$ is the input of GMM. $\omega_i$, $\mu_i$ and $\Sigma_i$ represents the weight, mean, and covariance matrix of the *i*-th Gaussian distribution, respectively. Moreover, $\sum \omega_i = 1$ and the probability distribution $N(X|\mu_i, \Sigma_i)$ is expressed as

$$N(X|\mu_i, \Sigma_i) = \frac{1}{2\pi^{d/2}|\Sigma_i|^{1/2}} exp\left\{-\frac{1}{2}(X-\mu_i)^T \Sigma_i^{-1}(X-\mu_i)\right\} \tag{14}$$

The parameters of the GMM are optimized using the Expectation-Maximization (EM) algorithm. This iterative optimization technique is employed to refine the GMM parameters, aiming to maximize the likelihood of the observed data. The EM algorithm operates through an iterative process involving two key steps: the Expectation step and the Maximization step.

In the Expectation step, the algorithm computes the posterior probabilities that each data point $x_i$ was generated by each Gaussian component $K$. These probabilities are denoted as

$$\Upsilon(z_{ik}) = \frac{\pi_k N(x_i|\mu_k, \Sigma_k)}{\sum_{j=1}^{K} \pi_j N(x_i|\mu_j, \Sigma_j)} \tag{15}$$

where $z_{ik}$ is the latent variable indicating the assignment of data point $x_i$ to component $K$, $N(x_i|\mu_k, \Sigma_k)$ is the probability density function of the Gaussian distribution with mean $\mu_k$ and covariance $\Sigma_k$.

Second, given the probabilities $\Upsilon(z_{ik})$ calculated in the Expectation step, the Maximization step updates the parameters $\theta = \{\omega_k, \mu_k, \Sigma_k\}$ to maximize the expected log-likelihood of the data by

$$\pi_k = \frac{1}{N} \sum_{i=1}^{N} \Upsilon(z_{ik}) \tag{16}$$

$$\mu_k = \frac{\sum_{i=1}^{N} \Upsilon(z_{ik}) x_i}{\sum_{i=1}^{N} \Upsilon(z_{ik})} \tag{17}$$

$$\Sigma_k = \frac{\sum_{i=1}^{N} \Upsilon(z_{ik})(x_i - \mu_k)(x_i - \mu_k)^T}{\sum_{i=1}^{N} \Upsilon(z_{ik})} \tag{18}$$

Consequently, the EM algorithm can ensure that the log-likelihood increases with each iteration, leading to convergence to a local maximum of the log-likelihood function. Specifically, the Expectation and Maximization steps are iteratively repeated until convergence, which is typically determined by the change in the log-likelihood $\log p(X|\theta) = \sum_{i=1}^{N} \log\left(\sum_{i=1}^{N} \pi_k N(x_i|\mu_k, \Sigma_k)\right)$ of the data between iterations falling below a predefined threshold.

## 1.3 Steam turbine unsupervised anomaly detection using ELSTMVAE-DAF-GMM

The steam turbine, a complex artificial system, consists of numerous components with diverse physical structures and relies on intricate sensors for monitoring and control. Due to various factors such as manufacturing variations, sensor quality, and operational conditions, the dataset collected from normal steam turbine operations inevitably contains certain anomalies and noise, significantly impacting detection performance. Moreover, for thermodynamic components of steam turbines, performance degradation caused by gradual factors may not manifest prominently in the short term, and anomalies often occur continuously, further complicating anomaly detection accuracy. Therefore, it is crucial to capture the temporal characteristics of steam turbine monitoring data over a period to ensure accurate anomaly detection. Additionally, the associated sensor data is remarkably intricate for components within a steam turbine, presenting considerable challenges in directly analyzing these high-dimensional datasets.

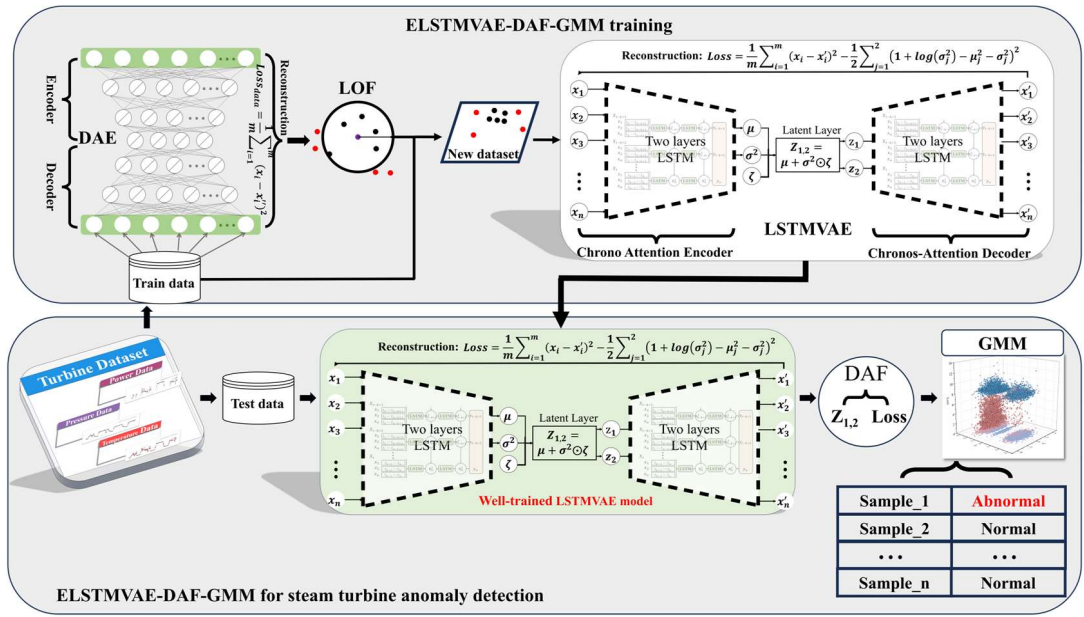

Fig. 4. Data flow through the present ELSTMVAE-DAF-GMM model including the training module, and anomaly detection module.

Based on the preceding analysis, data contamination, the temporal dynamic information, and the high-dimensional characteristics of steam turbine data remain three challenging tasks for achieving accurate steam turbine anomaly detection. While constructing anomaly detection algorithms directly for various types of steam turbine anomalies can mitigate the impact of data contamination, challenges arise due to the scarcity of well-labeled steam turbine anomaly samples and the costliness of data. To address this issue, a DAE-LOF sample selection mechanism is utilized to remove sample contamination from steam turbines, creating a new set of sample data. This dataset is then used to train the LSTMVAE model for feature learning, followed by unsupervised anomaly detection on steam turbines in a deep advanced feature space. Fig. 4 shows a detailed description of the process of performing unsupervised anomaly detection on steam turbines using the ELASTMVAE-DAF- GMM approach.

During the training process, a DAE-LOF sample selection algorithm was established to eliminate outliers in the normal data, enhancing the precision and reliability of anomaly detection results by reducing intrinsic outliers' influence. Specifically, we calculated the reconstruction errors of the training data samples using a trained Deep Autoencoder. Subsequently, the Local Outlier Factor algorithm was applied to identify and remove intrinsic noise and anomalies from the training samples based on the reconstruction error set. This process resulted in a refined training dataset composed of clean and healthy data. We constructed a chrono-attention reduced-order model, LSTMVAE, which was meticulously trained using this refined dataset. This model effectively maps the high-dimensional steam turbine data into a low-dimensional feature space. In the testing process, we computed deep advanced features for all steam turbine testing samples using the LSTMVAE model, which was well-trained on the refined training dataset. Finally, the deep advanced features were input into GMM for anomaly detection. To achieve robust overall performance as a primary objective for the ultimate detection task, we employed a diverse set of performance metrics to thoroughly evaluate the effectiveness of the proposed method. Algorithm 1 represents the training process of the ELSTMVAE-DAF-GMM.

**Algorithm 1: Training of the ELSTMVAE-DAF-GMM**

    Model DAE, encoder $E$ and decoder $D$ of the LSTMVAE.
    Input training data $X = \{x_1, x_2, \cdots, x_n\}$ is split into $X_{train}$ (first 80%), $X_{val}$ (remaining 20%).
    **repeat:**
        **for** $x$ **in** $X_{train}$:    # $x$ *represents different batches of samples.*
            $x' \leftarrow f_{DAE}(W_{DAE}x + b_{DAE})$
            **$Loss_{DAE}$** = MSE $(x, x')$
            **update** $W_{DAE}$ and $b_{DAE}$, representing the weight and bias matrices of the DAE neural network, respectively, by employing the backpropagation algorithm.
            **$Loss_{val}$** $\leftarrow$ computed from $X_{val}$
            **if** $Loss_{DAEval} < min(Loss_{DAEval}.history)$ **then**
                save model, $min(Loss_{DAEval}.history) \leftarrow Loss_{DAEval}$
            **else** *count* += 1
                **if** *count* > *stop_limit* **then** stop training    # *stop_limit is the stop setting*
                *of early stopping strategy.*
    **until** final epoch is reached
    $MSE_X \leftarrow$ trained model *DAE*, $X$
    $X_{refined} \leftarrow$ LOF ($MSE_X$, $C$)    # *LOF is the local outlier factor algorithm, C represents the contamination parameter.*
    Refined Dataset $X_{refined}$ is split into $X_{rtrain}$ (first 80%), $X_{rval}$ (remaining 20%).
    **repeat:**
        **for** $x_r$ **in** $X_{rtrain}$:    # $x$ *represents different batches of samples.*
            $\mu, \sigma \leftarrow \phi(x_r)$    # $\phi$ *is the function of the encoder in the LSTMVAE model.*
            $z \leftarrow \mu + \sigma^2 \odot \varepsilon, \ \varepsilon \leftarrow N(0, I)$
            $x'_r \leftarrow \Phi(z)$    # $\Phi$ *is the function of the decoder in the LSTMVAE model.*
            **$Loss = Loss_{rec} + KL$** $\leftarrow$ MSE $(x, x'), -\frac{1}{2}\sum_{j=1}^{2}(1 + log(\sigma_j^2) - \mu_j^2 - \sigma_j^2)$
            **update** the weight and bias of LSTMVAE using the backpropagation algorithm.
            **$Loss_{val}$** $\leftarrow$ computed from $X_{val}$
            **if** $Loss_{val} < min(Loss_{val}.history)$ **then**
                save model, $min(Loss_{val}.history) \leftarrow Loss_{val}$
            **else** *count* += 1
                **if** *count* > *stop_limit* **then** stop training    # *stop_limit is the stop setting of early stopping strategy.*
    **until** the final epoch is reached

## 1.4 Experiment data collection and descriptions

In this study, we conducted a comprehensive investigation into the real operational data of industry steam turbine equipment at a thermal power plant through a detailed case study. Through meticulous analysis of maintenance reports and monitoring data spanning the preceding two months, we aimed to provide valuable insights into the process of detecting anomalies, encompassing faults such as turbine blade wear. To ensure robust analysis, we performed essential data preprocessing on the raw operational data, which included database decoding, timestamp verification, alignment, and the removal of bad values (e.g., missing values and infinities) and faulty sensors. This meticulous preprocessing enabled us to curate a dataset comprising only periods with sufficiently complete data, addressing issues related to occasional data gaps due to sensor malfunctions.

Our primary aim is to identify and analyze anomalies caused by gradual degradation faults, particularly turbine blade wear, which, although subtle, can significantly impact efficiency and pose mechanical risks. Patterns of turbine blade wear faults typically include a gradual increase in vibration and friction, which, if left unaddressed, can lead to severe efficiency loss and potential mechanical failures. Timely and accurate detection of these faults is crucial, as undetected blade wear can escalate to significant mechanical issues, including unit vibration and friction problems within the turbine's flux parts.

As shown in Table 1, given the occasional unavailability of vibration data, the dataset includes data from 19 sensors that monitor key parameters such as temperature, pressure, and motor power across critical components like the turbine, main steam piping, and cylinders. These sensors were meticulously chosen to comprehensively monitor the operational state and performance of the steam turbine system. This selection of sensors is fundamental for effective anomaly detection, as it enables the identification of subtle variations and potential faults within the turbine's complex operational environment. The data collection spans from June 5, 2017, to July 17, 2017, with a one-minute sampling frequency per sensor. This dataset is divided into two distinct operational states: normal operation, spanning from June 5, 2017, to June 29, 2017, and abnormal operation, indicative of turbine blade wear, from July 13, 2017, to July 17, 2017. For model training, we exclusively used data from the normal operation phase. Specifically, data from June 5, 2017, to June 25, 2017, was used for training, with 20% of this subset reserved for validation. The remaining data from June 25, 2017, to June 29, 2017, was designated for testing.

Table 1. The sensors and descriptions for collected samples from industrial steam turbines.

| Number | Sensors | Descriptions |
|---|---|---|
| 1 | $P_0$ | Main steam piping main steam bus pressure #1 |
| 2 | $T_0$ | Main steam piping main steam bus temperature #1 |
| 3 | $P_1$ | Steam pressure in front of the main steam door on the right side of the main steam pipe #1 |
| 4 | $T_1$ | High-pressure valve #1 steam temperature in front of high-pressure main steam door |
| 5 | $P_2$ | High pressure valve regulating stage outlet steam pressure #1 |
| 6 | $T_2$ | High pressure valve regulating stage outlet steam temperature #1 |
| 7 | $P_3$ | High pressure valve high pressure exhaust steam pressure #1 |
| 8 | $T_3$ | High pressure valve high pressure exhaust steam temperature #1 |
| 9 | $P_4$ | Main steam piping medium pressure cylinder to low pressure cylinder connecting pipe steam pressure |
| 10 | $T_4$ | Medium pressure valve #2 steam temperature before reheating main steam door |
| 11 | $P_5$ | Medium pressure valve #2 pre-reheat main steam door pressure |
| 12 | $T_5$ | Low pressure cylinder inlet steam temperature |
| 13 | $P_6$ | Medium pressure valve #2 steam pressure after reheat regulator #1 |
| 14 | $T_6$ | Low pressure cylinder exhaust steam temperature #1 |
| 15 | $P_7$ | Medium pressure cylinder inlet pressure |
| 16 | $P_8$ | Medium pressure cylinder discharge pressure |
| 17 | $P_9$ | Low pressure cylinder water injection post regulator pressure |
| 18 | $P_{10}$ | Low pressure cylinder front connecting tube vapor pressure #1 |
| 19 | Eff | Generator active power #1 |

The performance of machine learning methods is heavily dependent on the properties of both the training and test data. It is crucial to assess how the distribution of test data differs from that of the training data. In this study, the training and test datasets represent similar operating conditions, ensuring a consistent basis for model evaluation. However, the potential for nominal changes in operating conditions to be detected as anomalies were also considered. While it is desirable for the model to detect significant deviations indicative of faults, detecting minor fluctuations that fall within the range of normal operational variability would lead to false positives. Therefore, careful attention was given to the selection and preprocessing of training data to encompass the typical range of operational conditions, thereby reducing the likelihood of false anomaly detections. This balance ensures the model's sensitivity to true anomalies, such as turbine blade wear, while maintaining robustness against nominal changes in operating conditions. By focusing on these periods and maintaining a consistent representation of operating conditions, we aim to enhance the reliability and efficiency of steam turbine systems, ensuring timely identification and mitigation of potential mechanical hazards. This approach contributes valuable insights to the field of industrial maintenance and operational optimization.

*1.5 Methods of evaluating classifier model performance*

In this study, we utilized the confusion matrix to assess the anomaly detection performance of our classifier model. This matrix serves as a fundamental metric in classification tasks, providing a detailed comparison between the model's predictions and the actual class labels of the samples. Table 2 presents the essential parameters of the confusion matrix: true positive (TP), false positive (FP), true negative (TN), and false negative (FN) values. The confusion matrix serves as the foundation for deriving essential performance metrics, including accuracy (AC), precision (PR), recall (RC), F1 score, and false alarm rate (FAR), which are widely utilized to evaluate the performance of anomaly detection algorithms. The computation formulas for these performance indicators are shown as follows. It's worth noting that higher values of AC, PR, RC, and F1 Score signify better model performance, while a lower FAR value is desirable to minimize the cost associated with misclassifications and enhance the overall effectiveness of the model.

$$Accuracy: AC = \frac{TN + TP}{TN + TP + FN + FP} \times 100\% \qquad (19)$$

$$Precision: PR = \frac{TP}{TP + FP} \times 100\% \qquad (20)$$

$$Recall: RC = \frac{TP}{TP + FN} \times 100\% \qquad (21)$$

$$F1\ Score: F1 = \frac{2 \times PR \times RC}{PR + RC} \times 100\% \qquad (22)$$

$$False\ alarm\ rate: FAR = \frac{FP}{FP + TN} \times 100\% \qquad (23)$$

Table 2. Confusion matrix.

|  |  | Predicted class | |
|---|---|---|---|
|  |  | Positive | Negative |
| Actual class | Positive | TP (true positive) | FP (false positive) |
|  | Negative | FN (false negative) | TN (true negative) |

## 2. Results and discussion

*2.1 Dimensionality reduction and anomaly detection performance of the ELSTMVAE-DAF-GMM model in industrial steam turbines*

To rigorously evaluate the performance of the proposed unsupervised anomaly detection model, this study utilized a set of key performance evaluation indicators, including accuracy (AC), precision (PR), recall (RC), F1 score, and false alarm rate (FAR). In order to comprehensively evaluate the performance of the proposed ELSTMVAE-DAF-GMM model, we conducted a series of experiments comparing various methods, contamination values, sequence lengths, and batch sizes, as well as ablation studies. The results, summarized in Table 3, demonstrate the superior performance and robustness of our approach across multiple metrics.

Our study involved a comparative analysis with four commonly used unsupervised anomaly detection methods, utilizing real operational data from steam turbines. Specifically, two of these methods are widely applied classical unsupervised algorithms (GMM [32], Kmeans [33]), while the other two are state-of-the-art deep learning-based methods (VAE-GMM [34], DAE-GMM [35]). Unlike the second and third methods, the fourth and fifth methods use Variational Autoencoders (VAE) and Deep Autoencoders (DAE) to extract features from raw data, followed by clustering using Gaussian Mixture Models (GMM). As depicted in Fig. 5, the proposed ELSTMVAE-DAF-GMM method leverages deep advanced features for anomaly detection, instead of relying solely on latent embeddings or reconstruction discrepancies. The DAF distinctly separates the latent space points of different modes, eliminating intersections and facilitating precise anomaly detection.

Table 3. Comprehensive results of anomaly detection performance, reliability analysis, and ablation studies of the ELSTMVAE-DAF-GMM model.

| No. | Method/Parameter | AC | PR | RC | F1 Score | FAR |
|---|---|---|---|---|---|---|
| Anomaly detection methods | ELSTMVAE-DAF-GMM | 94.6% | 94.9% | 94.6% | 94.6% | 5.43% |
| | GMM | 80.7% | 86.1% | 80.7% | 80.0% | 19.3% |
| | Kmeans | 81.5% | 86.5% | 81.5% | 80.9% | 18.5% |
| | VAE-GMM | 80.1% | 85.7% | 80.1% | 79.2% | 19.9% |
| | DAE-GMM | 80.2% | 85.8% | 80.2% | 79.4% | 19.8% |
| Contamination values | 0% | 91.9% | 92.9% | 91.9% | 91.9% | 8.07% |
| | 10% | 94.3% | 94.5% | 94.3% | 94.3% | 5.71% |
| | 20% | 94.6% | 94.9% | 94.6% | 94.6% | 5.43% |
| | 30% | 94.0% | 94.2% | 94.0% | 94.0% | 5.95% |
| Sequence length | 100 | 94.6% | 94.9% | 94.6% | 94.6% | 5.43% |
| | 150 | 94.0% | 94.2% | 94.0% | 94.0% | 6.03% |
| | 200 | 92.8% | 92.9% | 92.8% | 92.8% | 7.20% |
| | 250 | 93.5% | 93.8% | 93.5% | 94.0% | 6.47% |
| Batch sizes | 64 | 94.6% | 94.9% | 94.6% | 94.6% | 5.43% |
| | 128 | 93.9% | 94.2% | 93.9% | 93.9% | 6.05% |
| | 256 | 93.5% | 93.9% | 93.5% | 93.5% | 6.49% |
| | 512 | 92.0% | 92.1% | 92.0% | 92.0% | 7.98% |
| Ablation studies | ELSTMVAE-DAF-GMM | 94.6% | 94.9% | 94.6% | 94.6% | 5.43% |
| | ELSTMVAE-GMM | 74.6% | 83.2% | 74.6% | 72.9% | 25.4% |
| | LSTMVAE-DAF-GMM | 91.9% | 92.9% | 91.9% | 91.9% | 8.10% |
| | EVAE-DAF-GMM | 92.1% | 92.9% | 92.1% | 92.0% | 7.95% |

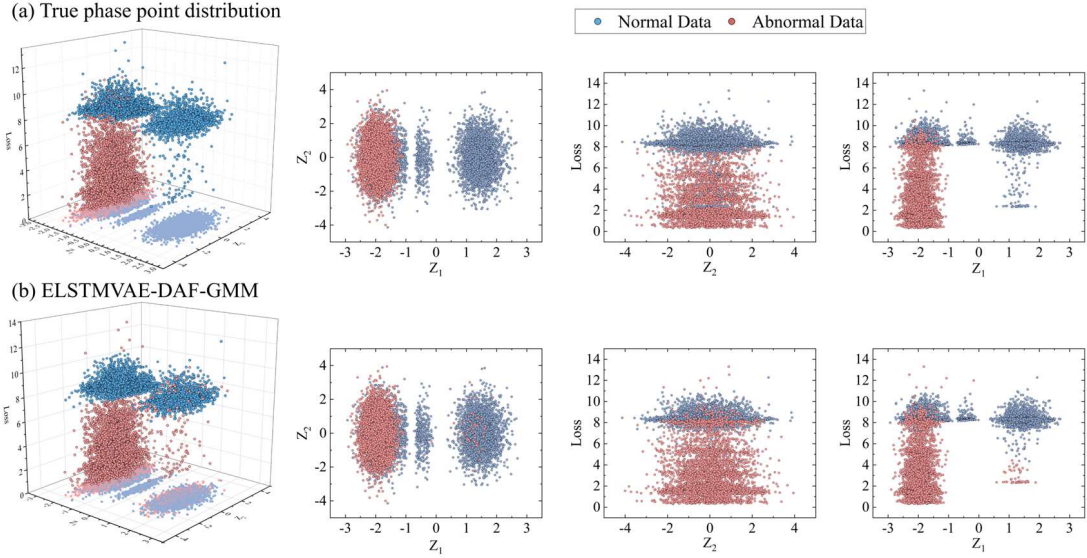

Fig. 5. Three-dimensional phase portraits of the deep advanced feature and their two-dimensional projections in the steam turbine dataset. (a) True phase point distribution. (b) ELSTMVAE-DAF-GMM clustering result. Different colors of phase points represent different classes of steam turbine data.

The training loss of the DAE and LSTMVAE within the proposed ELSTMVAE-DAF-GMM model is shown in Fig. 6. Throughout the training, both networks exhibit a steady decrease in both training and validation losses, indicating effective learning and generalization.

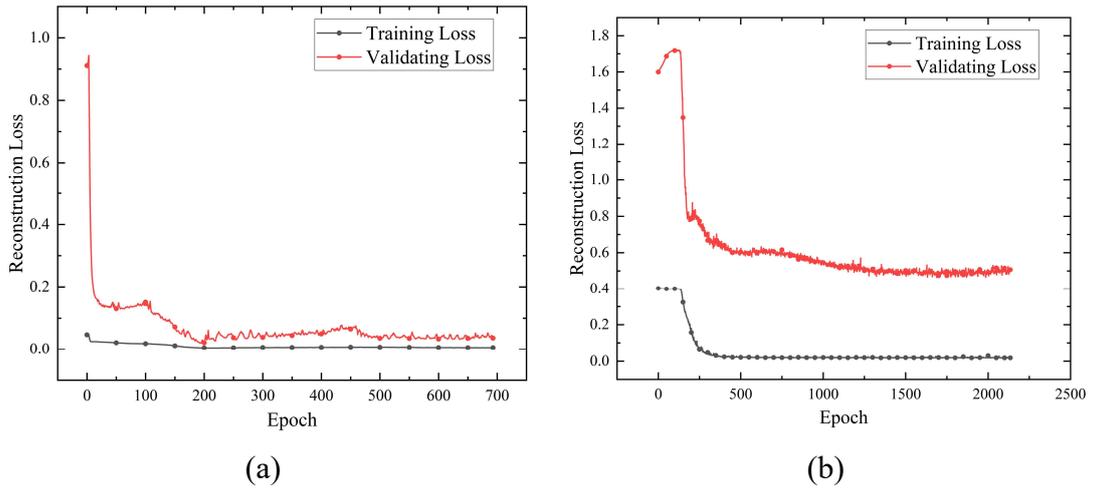

Fig. 6. Training loss of the DAE and LSTMVAE models within the proposed ELSTMVAE-DAF-GMM framework. (a) DAE training and validation loss. (b) LSTMVAE training and validation loss. The black curves represent training loss, while the red curves represent validation loss.

Fig. 7 presents the confusion matrix for the ELSTMVAE-DAF-GMM method compared with existing unsupervised methods in the steam turbine dataset. The confusion matrix provides detailed information about the model's predictive performance across different classes, allowing for further calculations to derive the above performance indicators.

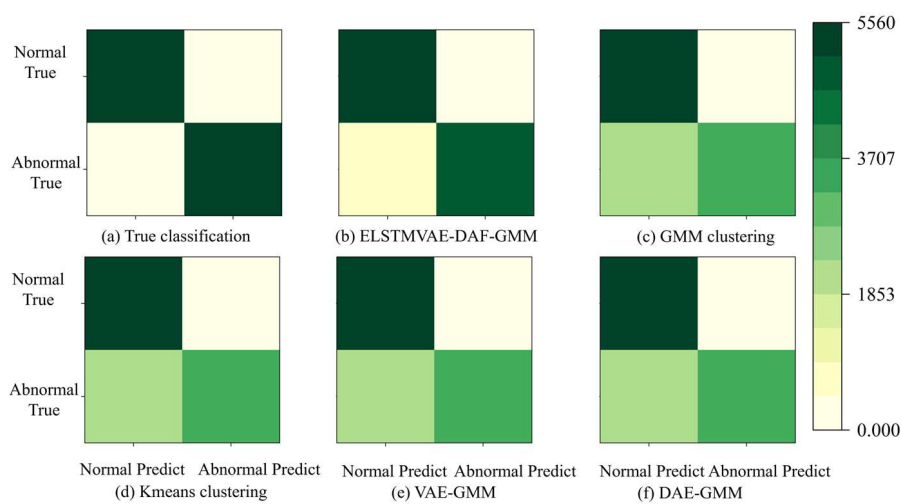

Fig. 7. Confusion matrix of the ELSTMVAE-DAF-GMM method compared with four existing unsupervised methods.

Fig. 8 illustrates the AC, PR, RC, F1 score, and FAR values of both the proposed method and four baseline methods. The ELSTMVAE-DAF-GMM method consistently achieves accuracy, precision, recall, and F1 score values exceeding 94.6%, indicating a substantial improvement over the other methods and significantly superior model performance. The false alarm rate (FAR) of 5.43% is notably lower than that of the other methods, suggesting a desirable reduction in misclassification costs and enhanced overall model effectiveness. These findings collectively attest to the outstanding performance of the proposed model among the compared methods. Specifically, the accuracy and recall values of ELSTMVAE-DAF-GMM surpass those of GMM, Kmeans, VAE-GMM, and DAE-GMM by up to 15.5%, demonstrating its superior ability to accurately identify both anomalies and normal samples, as well as to detect true anomalies. The precision value of ELSTMVAE-DAF-GMM exceeds those of the other methods by up to 9.9%, highlighting its accuracy in identifying abnormalities. Additionally, the F1 Score value of ELSTMVAE-DAF-GMM outperforms those of the other methods by up to 15.8%, elucidating its superior overall performance in steam turbine anomaly detection. The reduced FAR values, coupled with high positive metrics, reflect the effectiveness of ELSTMVAE-DAF-GMM in accurately identifying anomalies while minimizing false alarms, thus contributing to enhanced operational efficiency and reliability in steam turbine systems.

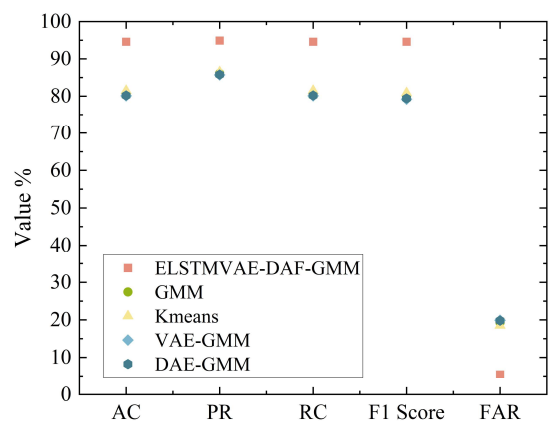

Fig. 8. Overall detection performance of the ELSTMVAE-DAF-GMM method compared with existing unsupervised methods using five performance metrics. Different types of phase points represent the results of different detection methods.

In addition to superior detection performance, the ELSTMVAE-DAF-GMM method also exhibits reasonable computational efficiency. As outlined in Table 4, although the ELSTMVAE-DAF-GMM method involves higher computational costs compared to the other four methods, its advantages in capturing complex temporal-spatial information of steam turbine system time-series data justify this trade-off. This approach still maintains a reasonable training time and a manageable number of parameters. This method excels in extracting intricate patterns from sequential data, making it particularly valuable for large-scale datasets. To further enhance its efficiency, various advanced adaptations of VAE and LSTM will be explored in Section 3.2, offering promising solutions for optimization. Although the complexity of the compared methods varies, the objective of this comparison is to focus on understanding how different model architectures impact anomaly detection performance in steam turbines.

Table 4. The computational cost of five approaches for the same steam turbine data.

| Methods | Training time (s) | Number of parameters |
|---|---|---|
| ELSTMVAE-DAF-GMM | 4405 | 7188 |
| GMM clustering | 1 | 420 |
| Kmeans clustering | 1 | 38 |
| VAE-GMM | 2829 | 771 |
| DAE-GMM | 1761 | 1135 |

In summary, the results underscore the robust performance of the proposed ELSTMVAE-DAF-GMM method, emphasizing its ability to achieve superior anomaly detection outcomes characterized by high positive indicators, minimal false alarm rates, and reasonable computational efficiency. This indicates the potential of the proposed approach to address the challenges associated with anomaly detection in complex steam turbine systems, contributing valuable insights to the field of industrial maintenance and operational optimization.

## 2.2 Reliability analysis of classification approaches

To ensure the robustness of our proposed methods, we performed an extensive sensitivity analysis on key parameters, including Contamination values (a crucial parameter of the DAE-LOF sample selection component), batch size, and sequence length (a crucial parameter of the two-layer LSTM neural network). Contamination values, such as 0%, 10%, 20%, and 30%, were varied systematically to assess the model's capability to filter out intrinsic noise and anomalies effectively. By adjusting the contamination level, we aimed to ensure that the filtered samples predominantly consisted of useless information, thereby refining the training set for better anomaly detection. Similarly, batch sizes, such as 64, 128, 256, and 512, were tested to identify the most effective size for optimizing training efficiency and detection accuracy. Additionally, we explored sequence lengths of 100, 150, 200, and 250 to evaluate the influence of temporal context on model performance. This comprehensive analysis aimed to fine-tune the ELSTMVAE-DAF-GMM method, ensuring its robustness and efficacy in detecting anomalies in steam turbine operational data.

Fig. 9 presents the anomaly detection results for steam turbine operation datasets at various contamination values of the DAE-LOF sample selection component. Our study identifies two key factors for optimizing anomaly detection algorithms: first, the removal of outliers during training enhances algorithm performance by allowing the model to focus on normal operational patterns, thus increasing precision and reliability. Second, selecting an appropriate contamination value is crucial. While metrics such as accuracy, precision, recall, and F1 score are relatively stable across different contamination values, the optimal performance is observed at 20%. Both excessively small and large contamination values can distort detection outcomes, either leaving residual anomalies or excessively removing original data, which inflates the false alarm rate (FAR). Therefore, it is essential to validate the filtering process by comparing performance metrics before and after filtering and to conduct controlled experiments with varying contamination values to prevent an inflated FAR or loss of important information.

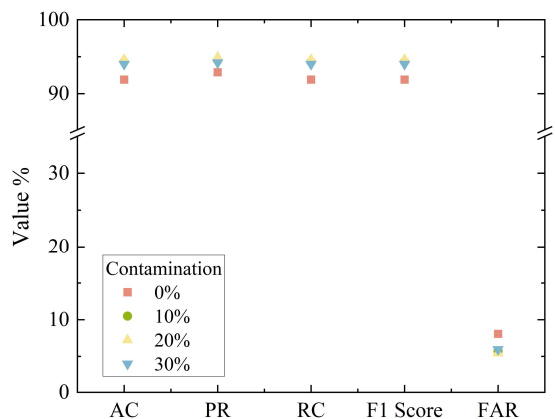

Fig. 9. Anomaly detection performance of the ELSTMVAE-DAF-GMM unsupervised anomaly detection approach across varying Contamination values in the steam turbine dataset.

Fig. 10 illustrates the sensitivity analysis of the ELSTMVAE-DAF-GMM method, assessing how sequence length and batch size impact anomaly detection performance in steam turbine operation datasets. The results reveal that as sequence length increased to 150, 200, and 250, and batch size increased to 128, 256, and 512, there were minor changes in performance metrics. Despite these variations, the algorithm consistently maintained high performance levels, demonstrating its robustness and stability across different parameter settings. Overall, the sensitivity analysis confirms that the ELSTMVAE-DAF-GMM method remains stable and reliable, with strong performance across a range of sequence lengths and batch sizes.

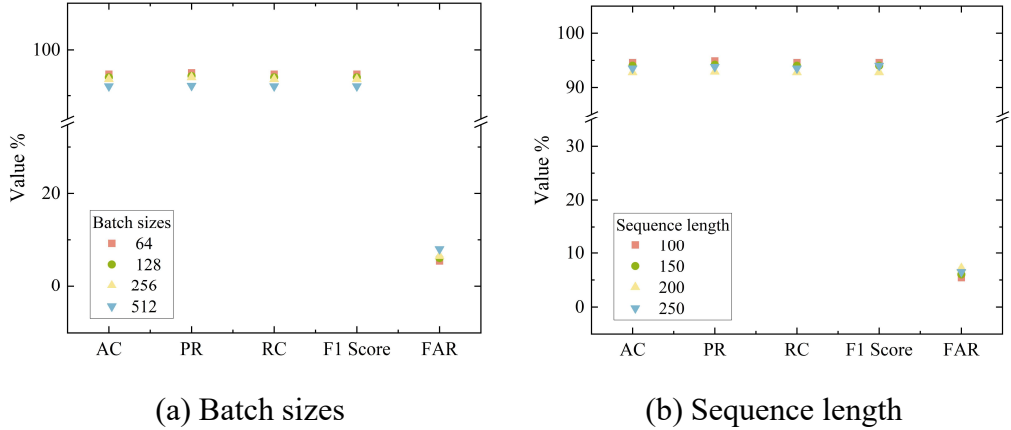

(a) Batch sizes  (b) Sequence length

Fig. 10. Sensitivity analysis of (a) batch sizes and (b) sequence length for the ELSTMVAE-DAF-GMM unsupervised anomaly detection approach for the steam turbine dataset.

To further evaluate the robustness of our proposed methods, we conducted additional case studies focusing on anomalies resulting from high-pressure cylinder regulating stage vapor seal wear failure in steam turbines, providing a comprehensive performance evaluation of our unsupervised anomaly detection methods. As illustrated in Fig. 11, the ELSTMVAE-DAF-GMM method consistently exhibited exceptional performance across a range of evaluation metrics, including accuracy (92.6%), precision (93.0%), recall (92.6%), and F1 score (92.6%), with a false alarm rate (FAR) of 7.42%. These results demonstrate the method's reliability in effectively distinguishing between normal and abnormal samples under varied fault conditions. The consistent performance across different types of anomalies confirms the robustness of our approach.

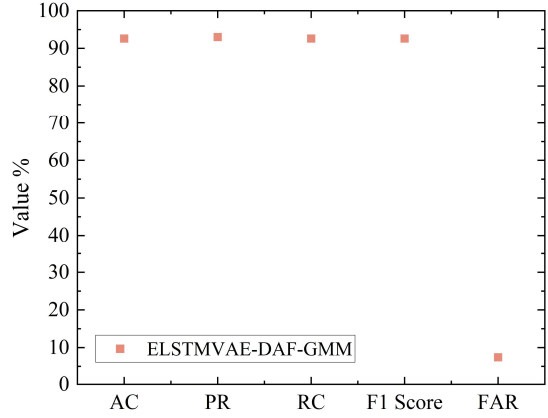

Fig. 11. Robustness analysis of the ELSTMVAE-DAF-GMM unsupervised anomaly detection methods for new case of anomalies resulting from high-pressure cylinder regulating stage vapor seal wear failure in steam turbines.

Careful management of LSTMVAE-based approaches for dimensionality reduction and classification is essential to avoid potential pitfalls. Improper model design or training can lead to posterior collapse, where the KL divergence term overwhelms the reconstruction term, limiting the latent space's capacity to capture meaningful data variations. Effective hyperparameter tuning, such as adjusting learning rate, batch size, sequence length, and neural network architecture, is essential to ensure the LSTMVAE converges to optimal solutions without sacrificing the diversity of learned latent representations. Variants like β-VAE [36] aim to mitigate posterior collapse but introduce additional complexity. Substituting LSTM with Gated Recurrent Units (GRUs [37]) can reduce computational complexity and speed up training but may compromise the model's ability to capture long-term dependencies. This necessitates careful consideration and tuning to balance performance and efficiency in practical applications.

## 2.3 Ablation studies

We conducted ablation studies to gain deeper insights into the individual contributions of the components comprising the proposed ELSTMVAE-DAF-GMM method. We systematically removed the DAE-LOF sample selection component, LSTMVAE component, and deep advanced feature component, to evaluate their impact on anomaly detection performance. By isolating these components, we aimed to discern their respective roles in enhancing the model's effectiveness and robustness.

As illustrated in Fig. 12, when the detection feature was modified to solely rely on latent variables, resulting in the ELSTMVAE-GMM approach that integrates the DAE-LOF sample selection module, the LSTMVAE module, and the GMM classifier, a marked decline in performance was observed. Specifically, there were significant reductions in accuracy, precision, and F1 score. This performance drop highlights the indispensable role of deep advanced features in improving the model's effectiveness, demonstrating that their inclusion is crucial for maintaining high detection accuracy and reliability. Further analysis involved removing the DAE-LOF sample selection component, resulting in the LSTMVAE-DAF-GMM method, which integrates the LSTMVAE module, deep advanced features, and the GMM classifier. Although this method still performed well, it exhibited decreased performance compared to the complete ELSTMVAE-DAF-GMM method. Notably, accuracy decreased from 94.6% to 91.9%, precision from 94.9% to 92.9%, recall from 94.6% to 91.9%, and F1 score from 94.6% to 91.9%, while the false alarm rate (FAR) increased from 5.43% to 8.10%. These changes highlight the importance of the DAE-LOF sample selection component in maintaining high model performance. Finally, removing the LSTM component to obtain the EVAE-DAF-GMM method, which includes the DAE-LOF sample selection module, the VAE module, deep advanced features, and the GMM classifier, led to further decreases in performance across all positive metrics but an increase in FAR, particularly in accuracy and precision. Specifically, accuracy decreased from 94.6% to 92.1%, and precision from 94.9% to 92.9%, reaffirming the significance of the LSTM component in extracting advanced features.

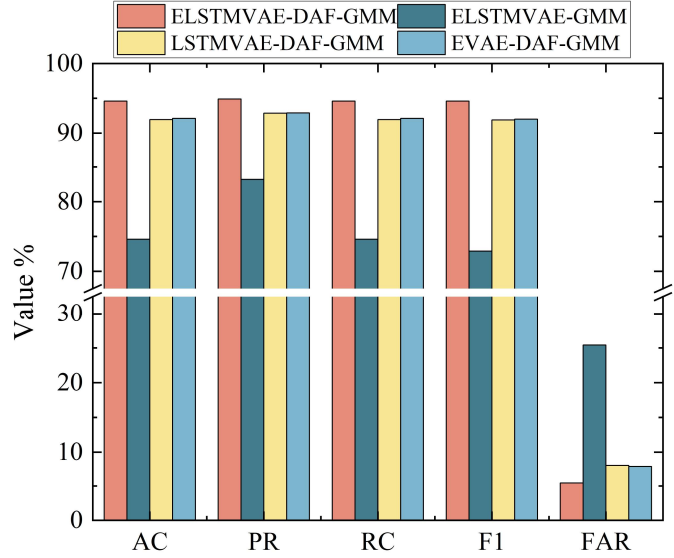

Fig. 12. Categorized performance of the ELSTMVAE-DAF-GMM clustering method, ELSTMVAE-GMM, LSTMVAE-DAF-GMM clustering method, and EVAE-DAF-GMM unsupervised anomaly detection method in steam turbine systems using five performance metrics. Different colors denote the anomaly detection results using different methods.

In summary, the results of the ablation studies clearly demonstrate the importance of each component in the ELSTMVAE-DAF-GMM method and provide specific numerical evidence to support these findings. This focused approach provides a deeper understanding of the underlying mechanisms driving the model's performance in steam turbine anomaly detection, thereby providing valuable insights for future research directions and further improvements.

## 3. Conclusions

Accurate anomaly detection is the prerequisite for ensuring the safe and stable operation of steam turbines. However, challenges such as inherent anomalies, insufficient temporal analysis, and the complexity of high-dimensional sensor data continue to limit the effectiveness of anomaly detection in steam turbines. To address these challenges, we proposed the Enhanced Long Short-Term Memory Variational Autoencoder using Deep Advanced Features and Gaussian Mixture Model (ELSTMVAE-DAF-GMM) for precise unsupervised anomaly detection in unlabeled steam turbine datasets. Compared with the conventional approaches, this method demonstrates superior performance, achieving high accuracy and minimal false alarm rates, making it a robust solution for the complexities inherent in steam turbine systems.

Firstly, the LSTMVAE network can effectively capture the temporal information of the complex steam turbine system, enabling prominent unsupervised anomaly detection performance in industrial steam turbine datasets. Secondly, the established DAE-LOF sample selection algorithm effectively eliminates inherent anomalous data points during training, enhancing precision and reliability by reducing outlier influence. This preprocessing step enables the model to focus on learning normal operational patterns, facilitating more effective identification and reporting of anomalies. Thirdly, a novel feature denoted as deep advanced features (DAF), combining latent features and reconstruction errors of LSTMVAE, is utilized to classify anomalies from unlabeled steam turbine data. This feature encompassed comprehensive information of the original high dimensional dataset and exhibited non-overlapping point distributions in the DAF phase space for different classes, consequently improving the overall performance of anomaly detection methods.

Our meticulously designed and rigorously analyzed approach demonstrates significant performance improvements in unsupervised anomaly detection for steam turbines. By addressing intrinsic anomalous data during training, capturing temporal information, and managing the high dimensionality of steam turbine datasets, our model enables accurate and proactive anomaly detection and maintenance strategies, ensuring the continuous safe and efficient operation of thermal power plants. Future work will focus on applying our methods to detect a wider range of turbine anomalies. Additionally, the demonstrated potential for analyzing timing data from industrial equipment suggests broader applicability beyond steam turbines.

**Declaration of competing interest**

The authors declare that they have no conflict of interest.

# Acknowledgments

This work is supported by the APRC-CityU New Research Initiatives/Infrastructure Support from Central of City University of Hong Kong (No. 9610601).

# Supplementary material

Supplementary material is available.